\documentclass[11pt,a4paper]{article}
\usepackage[hyperref]{emnlp-ijcnlp-2019}
\usepackage{cleveref}
\usepackage{times}
\usepackage{latexsym}
\usepackage{graphicx}
\usepackage{enumitem}
\usepackage{url}
\usepackage{tablefootnote}

\aclfinalcopy 

\definecolor{darkgreen}{rgb}{0.0, 0.2, 0.13}

\newcommand{\systemname}{{\sc CFO}}
\newcommand{\longsystemname}{{\sc Computation Flow Orchestrator}}

\title{\systemname{}: A Framework for Building Production NLP Systems}

\author{Rishav Chakravarti$^{1}$ \quad Cezar Pendus$^{2}$ \quad Andrzej Sakrajda$^{2}$ \quad Anthony Ferritto$^{1}$ \\
\textbf{Lin Pan$^{1}$  \quad Michael Glass$^{2}$ Vittorio Castelli$^{2}$ \quad J. William Murdock$^{1}$} \\
\textbf{Radu Florian$^{2}$ \quad Salim Roukos$^{2}$ \quad Avirup Sil$^{2}$\thanks{\quad Corresponding author.}}\\\\
$^{1}$IBM Watson, \quad $^{2}$IBM Research AI\\\\
\{rchakravarti, cpendus, ansa, panl, mrglass, vittorio, murdockj, raduf, roukos, avi\}@us.ibm.com \\
aferritto@ibm.com}

\date{}

\begin{document}
\maketitle
\begin{abstract}
This paper introduces a novel orchestration framework, called \systemname{} (\longsystemname{}), for building, experimenting with, and deploying interactive NLP (Natural Language Processing) and IR (Information Retrieval) systems to production environments. We then demonstrate a question answering system built using this framework which incorporates state-of-the-art BERT based MRC (Machine Reading Comprehension) with IR components to enable end-to-end answer retrieval.  Results from the demo system are shown to be high quality in both academic and industry domain specific settings. Finally, we discuss best practices when (pre-)training BERT based MRC models for production systems.
\end{abstract}

\section{Introduction}
\label{sec:intro}

Production NLP (Natural Language Processing) and IR (information retrieval) applications
often rely on a system flow consisting of multiple components that need to be woven together
to build an end-to-end system \cite{ferrucci2010building, yang2019endtoend}. This paper presents a novel approach for defining flow graphs and a toolkit for compiling those definitions into deploy-able production grade systems\footnote{The toolkit is available at \url{http://ibm.biz/cfo_framework}}.  

Though the framework, which we refer to as \systemname{} (\longsystemname{}), is well suited to a variety of use cases, we demonstrate it by creating an interactive QA (Question Answering) system that can be used both for academic benchmarking as well as industry specific use cases.  The interactive system integrates SOTA (state-of-the-art) BERT-based MRC (Machine Reading Comprehension), an Elasticsearch based document retrieval component, and a de-duplication \& sorting component to provide end-to-end answer retrieval.  We will refer to this demonstration system as GAAMA (Go Ahead, Ask Me Anything). The key contributions of this work, therefore, are to (1) introduce a novel framework for stitching together deployable NLP components, (2) demonstrate the framework with an end-to-end QA system, and (3) discuss the training steps necessary for adapting a SOTA MRC model to a data set before plugging it into the QA system.

\Cref{sec:system-arch} provides the motivations and details of the \systemname{} 
framework, \cref{sec:model} discusses the specific model components 
integrated into GAAMA, \cref{sec:experiment} discusses experimentation to adapt the BERT-based MRC component to this system, \cref{sec:relatedwork} discusses related work, and, finally, \cref{sec:conclusion} provides a conclusion and discussion of future work.

In addition, a (private) screencast video demonstration of GAAMA has been uploaded at \url{http://ibm.biz/gaama_demo} (along with a supplementary presentation of the CFO framework at \url{http://ibm.biz/gaama_cfo_demo}).
\section{\systemname{} Architecture}
\label{sec:system-arch}

The CFO framework relies on two sets of system specifications to define the computation flow graph.  First, each node within the graph defines its service name, input message data fields, and output message data fields using Google's Protocol Buffer Interface Definition Language\footnote{\url{https://developers.google.com/protocol-buffers/}}. Second, the orchestrator is described using a custom specification format allowing the user to declare:
\begin{enumerate}[nolistsep]
\item The set of nodes (provided as containerized gRPC\footnote{\url{https://grpc.io/}} microservices). Each node will have implemented a microservice according to the node's declared interface specification.
\item The set of nodes to treat as entry points to the flow graph. 
\item A mapping within the flow for each \textit{data element} comprising the input and output message interfaces for nodes. 
\item For ease of deployment, the specification also provides the ability to describe deployment specific configuration settings like service ports, docker registry location etc.
\end{enumerate}

Given these specification files, CFO provides a compiler to auto-generate an orchestrator node which implements the computation flow graph, a (dockerized) launch script, and a simple REST interface / GUI which provide access to the defined entry points and debug information.

The resulting design allows data flows and dependencies to be described both concisely and transparently.  The flow specification allows for easy debugging and modification of how data fields enter the computation flow, get transformed, and finally outputted. This is a significant departure from traditional orchestration systems which would require parsing through source code to determine and change the route taken by data fields through the computation flow. Furthermore, the use of Protocol Buffers to describe these data fields ensures a language and platform agnostic representation that does not compromise the system's speed or ability to ensure data type correctness at compile time (as opposed to traditional JSON/XML representations which need to encode/decode from strings at run time).

Another benefit of the CFO toolkit is to auto-generate the serialization and connectivity code for each node as well as exposing the entry points via REST interfaces.  Relinquishing this responsibility to the CFO toolkit frees the developer up from writing a significant amount of boilerplate code and worrying about distributed system best practices such as enforcing time outs, error propagation, latency logging, and parallelization through asynchronous calls. As an illustrative example, the auto-generated code for the GAAMA demo consists of 2,800 lines of C++ orchestration code.

Finally, the CFO toolkit generates a set of shell scripts, docker images, and config files for deploying and running the flow graph using either docker-compose\footnote{\url{https://docs.docker.com/compose/}} or kubernetes\footnote{\url{https://kubernetes.io/}}.  This allows the generated project to be deployed both locally for debugging as well as on modern cloud infrastructure.
\section{GAAMA Architecture}
\label{sec:model}

\begin{figure}
\begin{center}
\fbox{\includegraphics[trim = 20mm 15mm 25mm 2mm, clip,width=\columnwidth]{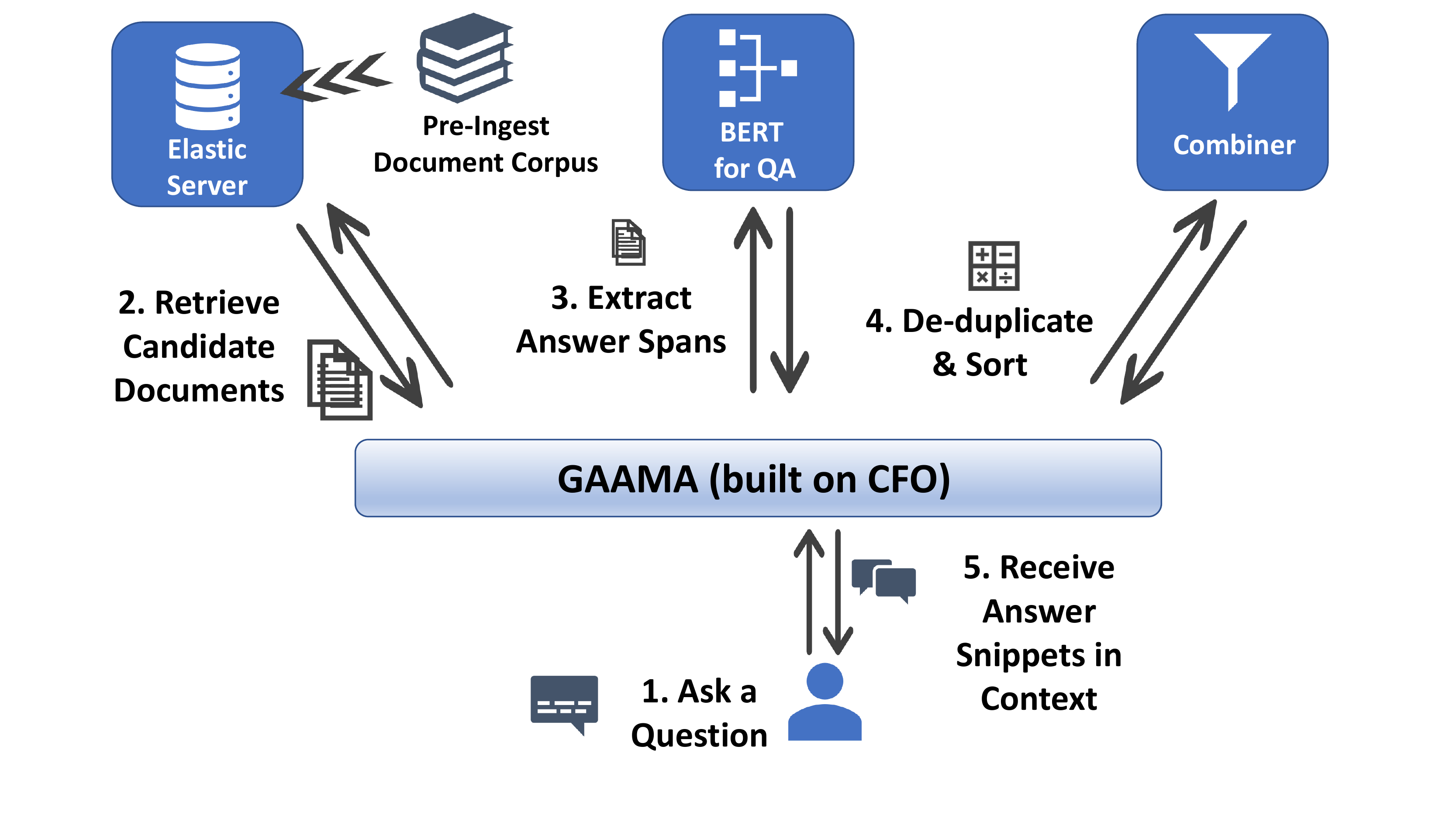}}
\end{center}
\caption{GAAMA System Architecture}
\label{fig:gaama}
\vspace{-1.5em}
\end{figure}

As a simple case study for \systemname{}, we create a demonstration QA system consisting of four nodes: 
(1) an Elasticsearch\footnote{\url{https://www.elastic.co/products/elasticsearch}} based IR node (2) a BERT based MRC node (3) an answer ``de-duplication'' node and (4) a final answer combiner node.  See \cref{fig:gaama} for an overview of the QA system.

As described in \cref{sec:system-arch}, each component starts by declaring an interface specification allowing the underlying implementation to be swapped out without the need to modify the rest of the QA system. Next, a gRPC server is implemented with the core business logic.  Auto-generated gRPC code stubs provide the core communication/serialization logic for the service layer of the server. We describe the IR and MRC nodes in further detail in the following two sections.

\subsection{IR with Elasticsearch}
The core business logic of the IR node uses Elasticsearch APIs to retrieve the top $k$ documents from the appropriate corpus based on BM25, a popular variant of term frequency overlap between the query and document text \cite{Robertson_theprobabilistic}. Two document corpora are ingested using the standard English analyzer. The first corpus consists of  Wikipedia paragraphs used for academic benchmarking and the second corpus consists of an industry dataset made up of IBM Technical Support Documentation. The user interface allows us to choose either corpus when asking questions to evaluate the system. We use ``paragraphs'' as the base unit of ingestion in line with \cite{yang2019endtoend} which shows this as an optimal pre-processing step for consumption by a MRC component.

The node's input interface, therefore, accepts query text, a hyperparameter $k$, and a target corpus identifier. Its output interface produces a list of document texts accompanied by retrieval scores. As discussed earlier, these interfaces are defined using protocol buffer definitions, so we follow standard steps\footnote{\url{https://grpc.io/docs/tutorials/}} for auto-generating a gRPC server with placeholders for the custom business logic of retrieving documents. Similarly, with the server logic in place, we follow standard steps\footnote{\url{https://github.com/grpc/grpc-docker-library/}} to create a docker image that \systemname{} will need to launch the gRPC server. Though we are wrapping Elasticsearch's index based retrieval implementation here, we can swap our implementation for more recent Neural IR based techniques \cite{craswell2017neuir} without changing the exposed interface or the orchestration code.

\subsection{MRC with BERT}
\label{subsec:bertmodel}
The MRC node similarly wraps a BERT-for-QA model in a dockerized gRPC server which accepts a single query-document pair as its input and produces a span from the document along with a prediction score as its output. Note that \systemname{}'s orchestrator automatically realizes that the IR node produces a list of documents, while the MRC node accepts a single document at a time. So \systemname{} will take care of calling the MRC node for each of the $k$ retrieved documents (using asynchronous calls to parallelize requests if a configuration flag is set). 

The underlying BERT-for-QA model is based on \cite{alberti2019bert}. BERT \cite{Devlin2018BERTPO} is one of a series of pre-trained neural models that can be fine tuned to provide state-of-the-art results in NLP \cite{Peters_2018,howard2018universal,radford2019opengpt} including on the SQuAD \cite{rajpurkar2018know} and NQ \cite{Kwiatkowski2019NaturalQA} tasks that align with our MRC based QA.
 
 \begin{figure}
\includegraphics[trim = 2mm 2mm 2mm 2mm, clip,width=1\columnwidth]{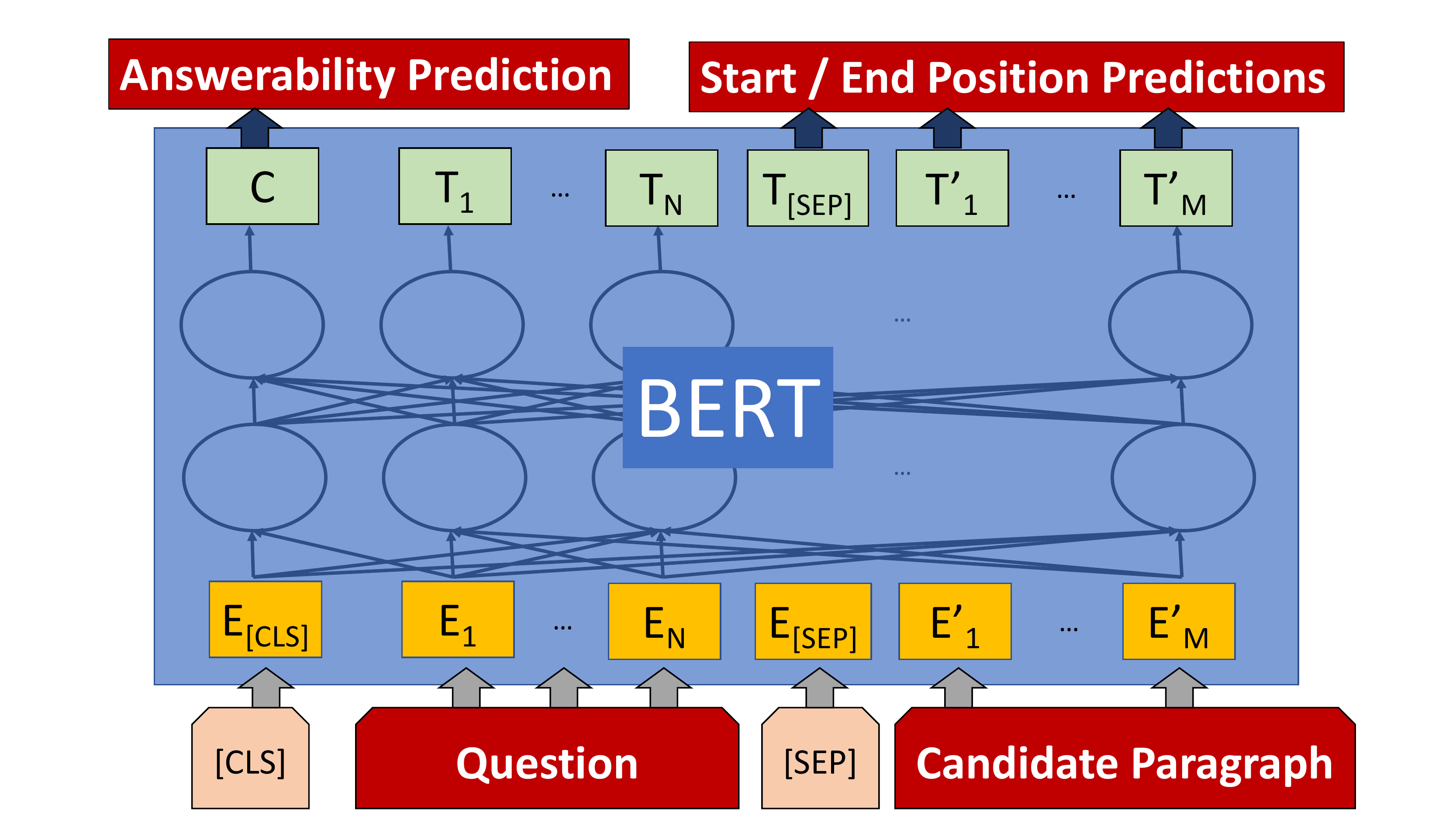}
\caption{BERT for QA \cite{Devlin2018BERTPO}}
\label{fig:bert}
\vspace{-1.5em}
\end{figure}
 
We use the Huggingface PyTorch implementation of BERT \footnote{\url{https://github.com/huggingface/pytorch-transformers}} which supports starting from a Base (a 12 layer, 768 hidden dimension, 12 attention head, 110M parameter transformer network) or a Large (a 24 layer, 1024 hidden dimension, 16 attention head, 340M parameter transformer network) model. An output feed forward layer is added on top of this to produce 3 sets of scores: (1) scores at each token offset marking the likelihood of an answer chunk starting at this offset (2) scores at each token offset marking the likelihood of an answer chunk ending at this offset (3) a score for the entire sequence marking the likelihood of the question being answerable given the current context. 


The parameters of the entire network are fine tuned using a set of question and document pairs where annotators provide the correct start and end offsets or have marked the question-document pair as having no correct answer. Refer to \cref{fig:bert} for a visual depiction of the model and to \cite{alberti2019bert} for additional details about model architecture and implementation.

\Cref{sec:experiment} includes practical pre-training, fine tuning, and hyperparameter optimization steps for building the final model deployed as part of GAAMA. At the time of writing, our BERT-based MRC model\footnote{The best performing submission to the leaderboard uses an ensemble of models using different hyperparameters rather than a single model} was the best performing submission  (dated 7/31/2019), outperforming the next best system by 1\% on F1 on the Natural Questions public leaderboard\footnote{\url{https://ai.google.com/research/NaturalQuestions/leaderboard}}.
 
\section{Experiments}
\label{sec:experiment}
Prior to integration of the MRC node into GAAMA, we first experiment with data preparation and training of the BERT MRC model as a standalone component using dev sets provided with the NQ and SQuAD data sets (see \ref{sec:relatedwork} for more on these data sets). NQ is preferred for evaluating production systems since the questions were ``naturally'' generated and does not suffer from the observational bias inherent in SQuAD's data collection approach \cite{Kwiatkowski2019NaturalQA}. When reporting results with the SQuAD dataset, we use the methodology (and evaluation script) made available with \cite{rajpurkar2018know}. Similarly, when reporting results with the NQ dataset, we use the methodology (and evaluation script) made available with \cite{Kwiatkowski2019NaturalQA}.  Once we are satisfied with the performance of this model, we integrate into GAAMA and evaluate manually using an internal corpus.

We use the F-score at an ``optimal'' threshold for the dev set\footnote{See \url{http://www.ibm.biz/confidence_thresholding} for more on choosing business specific thresholds} as the headline metric for assessing the system.  Latency measurements are carried out using a random sample of examples on a system with an Intel\textsuperscript{\textregistered} Xeon\textsuperscript{\textregistered} E5-2690 16-core CPU, 2 Nvidia\textsuperscript{\textregistered} Tesla\textsuperscript{\textregistered} P100 GPUs, and 128GB of RAM.\footnote{We only use 1 P100 GPU or 8 CPU threads in latency experiments}  We then examine the feasibility of deploying base and large models in a production environment on GPUs and CPUs.

\begin{table}[]
\scalebox{.9}{\begin{tabular}{lll}
\hline
& \textbf{F1} \\ \hline \hline
\textbf{Prior Work} \\ \hline \hline
DecAtt + Doc Reader \cite{Parikh_2016} & 31.4 \\ \hline
BERT \cite{Devlin2018BERTPO} & 50.2 \\ \hline
BERT w/ SQuAD 1.1 \cite{alberti2019bert} & 52.7 \\ \hline \hline
\textbf{This Work} \\ \hline \hline 
BERT w/ U-MRC & 53.6 \\ \hline
BERT w/ U-MRC \& SQuAD 1.1 & 54.2 \\ \hline
BERT w/ U-MRC \& SQuAD 1.1 \\ + SQuAD 2.0 Data Aug & \textbf{54.5} \\ \hline
\end{tabular}}
\caption{Dev Set Performance on NQ with different pre-training \& data augmentation techniques. We also report some baselines from \cite{alberti2019bert} for context }
\label{tab:pretraining}
\vspace{-1.5em}
\end{table}

\subsection{Pre-Training \& Data Augmentation}

We explore two types of pre-training.  The first follows \cite{alberti2019bert} by leveraging a similar task for which supervised labels are available and pre-training the model on it before moving onto fine tuning on the target dataset. Specifically, we use SQuAD 1.1 \cite{Rajpurkar_2016}.  \Cref{tab:pretraining} shows that this strategy can provide an absolute improvement of 2.5\%  over a model that starts with just the default BERT language model. 

We also employ \cite{Anon_2019}'s approach to using an unsupervised auxilary task that is better aligned to our final task (i.e. MRC) than the default Masked Language Model and Next Sentence Prediction used in \cite{Devlin2018BERTPO} to pre-train the BERT models.  Using the Wikipedia corpus, we create cloze style queries by masking out terms (named entities or noun phrases) in a sentence.  Then we identify an answer bearing passage from the Wikipedia corpus that is relevant to the query (as identified by BM25 IR).  This allows us to pre-train all layers of the BERT model including the answer extraction weights by training the model to extract the answer term from the selected passage.  Like the Masked Language Model, this task relies on predicting a masked component of an input sequence, but the prediction is generated by extraction rather than generation. \Cref{tab:pretraining} labels these results as ``BERT w/ U-MRC'' and shows that this additional training on a MRC specific unsupervised task improves the model's final fine-tuned performance on the NQ task by 1.5\%. \Cref{tab:pretraining-squad} similarly shows the benefits of these pre-training strategies on the SQuAD 2.0 dataset.

In addition, as noted by the authors of the original BERT-for-QA submission to SQuAD \cite{Devlin2018BERTPO}, there can be a benefit to fine tuning the entire network with labelled examples from multiple datasets. The last row of \cref{tab:pretraining} shows an incremental gain of 0.3\% by introducing SQuAD 2.0 during the fine-tuning phase. For now, the additional data is simply shuffled into the first 80\% of mini batches during the fine-tuning phase.

\begin{table}[]
\begin{center}
\scalebox{.9}{\begin{tabular}{lll}
\hline
Pre-Training & EM & F1 \\ \hline \hline
BERT \cite{Devlin2018BERTPO} & 78.7 & 81.9 \\ \hline
BERT w/ U-MRC & 82.2 & 85 \\ \hline
BERT w/ U-MRC \& NQ & \textbf{82.6} & \textbf{85.4} \\ \hline
\end{tabular}}
\caption{Dev Set Performance on SQuAD 2 with different pre-training strategies.}
\label{tab:pretraining-squad}
\end{center}
\end{table}

\subsection{BERT Models \& Latency}
Most model settings are taken from \cite{alberti2019bert} with the exception of batch size and learning rate which are tuned using the approach from \cite{smith2018disciplined}. In addition, we experiment with  models trained on BERT base and BERT large to understand trade-offs between latency and accuracy.  Using the hardware described in \cref{sec:experiment}, we evaluate F1 and latency on a subset of the NQ dev set\footnote{Used 500 random examples from dev set for experiments}.  In order to decrease latency, we simulate passage retrieval to send GAAMA the most relevant passage by selecting the first correct top level candidate if there is one and the first (incorrect) top level candidate if there is not.  We find in \cref{tab:f1-latency-hyperparam} that switching from base to large yields an 8.3\% absolute increase in F1 in exchange for 1.3x to 2.8x increases in latency.  When running GAAMA on a GPU these result in manageable 95th percentile query latencies of less than a second; whereas on the CPU the 95th percentile times are in excess of two and five seconds for base and large respectively.  For large, even the median latency is greater than one and a half seconds, effectively cementing GPUs as a requirement for deploying to production environments. In future work we intend to explore network pruning or knowledge distillation techniques for potential speedups with the large model.

\begin{table}[]
\centering
\scalebox{1}{
\begin{tabular}{lllllll}
\hline
Model & $F1$ & $T^{G}_{50}$ & $T^{G}_{95}$ & $T^{C}_{50}$ & $T^{C}_{95}$ \\ \hline \hline
Base & 42.5 & 0.05 & 0.49 & 0.53 & 2.32 \\ \hline
Large & 50.8 & 0.10 & 0.66 & 1.51 & 6.00 \\ \hline
\end{tabular}
}
\caption{F1 and latencies for BERT base and large models running on GPU and CPU for a subset of the NQ dev set.  $T^{D}_{K}$ is the $K$-th percentile query latency in seconds when running on device $D$ (GPU or CPU).}
\label{tab:f1-latency-hyperparam}
\vspace{-1em}
\end{table}

\section{Related Work}
\label{sec:relatedwork}

Recently \cite{yang2019endtoend} proposed BERTSerini, an end-to-end QA pipeline demo that leverages the Anserini IR toolkit \cite{Yang:2017:AEU:3077136.3080721} to look for relevant documents for a question, then uses BERT-based techniques \cite{Devlin2018BERTPO} to extract the correct answer.  However, their reliance on a Lucene based IR toolkit means that constructing a NLP pipeline would either require pipeline components to be written as Lucene based plugins (which comes with a variety of constraints on programming language and structure) or writing custom orchestration code to connect components outside of the toolkit. Similar constraints are imposed by other popular NLP pipeline toolkits such as StanfordNLP \cite{qi2018universal} and Spacy\footnote{\url{https://spacy.io/}} (both of which require development in Python with limited flexibility in training neural models with other frameworks such as Tensorflow\footnote{\url{https://www.tensorflow.org/}}).  

In contrast, \systemname{}'s inherent programming language and platform agnostic microservice architecture encourages flexibility and robustness in being able to switch out individual components without re-doing boilerplate code.  In addition, \systemname{}'s out-of-the-box support for containerization provides flexibility in the compute infrastructure that can be leveraged for rapid deployment to both local and cloud environments.

This flexibility is important in a domain such as Machine Reading Comprehension (MRC) where recent advances in language-modeling based pre-trained embeddings like ELMO \cite{Peters_2018} and BERT \cite{Devlin2018BERTPO} along with large scale open data sets like the Stanford Question Answering Dataset (SQuAD) 1.1 \cite{Rajpurkar_2016} and its successor SQuAD 2.0 \cite{rajpurkar2018know} have spurred a diverse array of model architecture improvements in a short time span. Recent work has even produced systems that surpass human-level exact match accuracy on the SQuAD datasets, causing us to focus on the challenging new Natural Questions (NQ) dataset \cite{Kwiatkowski2019NaturalQA} where the questions do not have any observational bias as they were not artificially created. To the best of our knowledge, there is no current software framework paper that shows its analysis on the NQ dataset and displays strong empirical performance.


UIMA \cite{ferrucci_lally_2004} is an integration framework that provides defined APIs for analyzing unstructured information and a shared data structure for storing the results of that analysis.  When paired with the UIMA Asynchronous Scaleout layer \cite{uimaas2018} and the Distributed UIMA Cluster Computing tool \cite{ducc2013}, this technology stack provides many of the same core capabilities that \systemname\ does: pipeline orchestration, microservice deployment and management, data serialization and connectivity, etc.  However, \systemname\ is designed for modern cloud environments and includes built-in integration with docker-compose and kubernetes; the UIMA stack \emph{can} be used with these technologies but facilities for doing so are not built-in to the stack so more development effort is needed in those contexts.  Also, UIMA and \systemname\ both require that each component expresses its data model (including input and output specifications) declaratively, but UIMA then unifies the data model of all components into a global type hierarchy, which requires some level of compatibility across type definitions.  In contrast, \systemname\ only requires consistency in the data model for components that directly connect to each other, and the names of corresponding types and fields do not need to match.  These differences make \systemname\ easier to use in cases where components were developed by different developers and integrated by a third party.



\section{Conclusion}
\label{sec:conclusion}
This paper introduces \systemname{}, a novel methodology and toolkit for rapid development of production grade systems for use cases which can be represented as computation flow graphs. We demonstrate the use of this framework to build an end-to-end QA system composed of a SOTA MRC model that is adapted to answering ``natural language questions''.  We also show experimentation using the NQ dataset to illustrate training techniques that can be used to build SOTA systems on top of existing pre-trained language models like BERT.

We are actively seeking to open source the \systemname{} framework and hope that, once available, the community will be able to quickly build and deploy their own SOTA NLP components as interactive multi-component systems.  We also intend on expanding GAAMA to incorporate additional QA components to improve its performance through approaches like query expansion for improved recall and network pruning for improved latency.

\bibliography{gaama}
\bibliographystyle{acl_natbib}

\end{document}